\documentclass[runningheads]{llncs}
\usepackage[T1]{fontenc}
\usepackage{graphicx}
\usepackage{booktabs}
\usepackage[misc]{ifsym}
\usepackage{multirow}
\usepackage{amsmath}
\usepackage{hyperref}
\usepackage{amssymb}

\newcommand{\corr}{(\Letter)}
\usepackage{mwe}

\begin{document}

\title{Enabling Granular Subgroup Level Model Evaluations by Generating Synthetic Medical Time Series}

\titlerunning{Synthetic ICU Time Series for Robust Model Evaluation}

\author{
Mahmoud Ibrahim\inst{1,2,3}\corr \and
Bart Elen\inst{3} \and
Chang Sun\inst{1,2} \and
Gökhan Ertaylan\inst{3} \and
Michel Dumontier\inst{1,2}
}

\authorrunning{M. Ibrahim et al.}

\institute{
Institute of Data Science, Faculty of Science and Engineering, Maastricht University, Maastricht, The Netherlands
\and
Department of Advanced Computing Sciences, Faculty of Science and Engineering, Maastricht University, Maastricht, The Netherlands
\and
VITO, Belgium\\
\email{mahmoud.ibrahim@vito.be}
}

\maketitle              

\begin{abstract}
We present a novel framework for leveraging synthetic ICU time‐series data not only to train but also to rigorously and trustworthily evaluate predictive models, both at the population level and within fine‐grained demographic subgroups. Building on prior diffusion and VAE‐based generators (TimeDiff, HealthGen, TimeAutoDiff), we introduce Enhanced TimeAutoDiff, which augments the latent diffusion objective with distribution‐alignment penalties . We extensively benchmark all models on MIMIC‐III and eICU, on 24‐hour mortality and binary length‐of‐stay tasks. Our results show that Enhanced TimeAutoDiff reduces the gap between real‐on‐synthetic and real‐on‐real evaluation (“TRTS gap”) by over 70 \%, achieving $\Delta_{TRTS} \leq $ 0.014 AUROC, while preserving training utility ($\Delta_{TSTR} \approx $  0.01). Crucially, for 32 intersectional subgroups, large synthetic cohorts cut subgroup‐level AUROC estimation error by up to 50 \% relative to small real test sets, and outperform them in 72–84 \% of subgroups. This work provides a practical, privacy‐preserving roadmap for trustworthy, granular model evaluation in critical care, enabling robust reliable performance analysis across diverse patient populations without exposing sensitive EHR data, contributing to the overall trustworthiness of Medical AI.

\end{abstract}

\section{Introduction}

Machine learning in critical care increasingly depends on large‐scale ICU time‐series data (continuous and irregular measurements of vitals, labs, and interventions) to develop predictive tools such as early‐warning systems, mortality risk estimators, and length‐of‐stay predictors. Despite extensive repositories like eICU \cite{eicu} and MIMIC \cite{mimic}, data sharing remains constrained by privacy regulations and limited access for underrepresented populations. Synthetic ICU time‐series data offer a promising solution by providing realistic, privacy‐preserving alternatives to real patient records. Recent advances in generative models, from GANs \cite{EHR-M-GAN} to diffusion frameworks \cite{TimeDiff}, have shown strong fidelity in reproducing complex temporal patterns of physiological data.

Beyond enabling model training without privacy exposure, synthetic data can also support evaluation when access to real test data is restricted. For instance, a model trained on institutional EHRs may need to be evaluated externally by regulators or collaborators, without sharing the real data. In such cases, a synthetic hold‐out set should faithfully approximate the real test distribution so that performance metrics (e.g., AUROC, RMSE) on synthetic samples reflect true performance on real patients.
Most prior work, however, emphasizes training utility: the “Train on Synthetic, Test on Real” (TSTR) setup while neglecting evaluation utility, or “Train on Real, Test on Synthetic” (TRTS). For synthetic data to be a reliable test substitute, models evaluated on synthetic hold‐outs must achieve performance comparable to those tested on genuine real data.

Beyond global model evaluation, a third, and often overlooked use case for synthetic ICU data is subgroup‐level evaluation. ICU cohorts are inherently heterogeneous: patients differ in age, gender, ethnicity, comorbidities, socioeconomic status, and other factors. A high‐impact clinical question is “How does my mortality prediction model perform on patients aged over 75, or on Black patients, or on females, or on black females aged over 75?” In this work, the focus is on proper model evaluation, specifically on addressing algorithmic bias through performance analysis on diverse population subgroups, which is recommended by ~\cite{subgroup-evaluation}. Unfortunately, real EHR datasets frequently contain only a few hundred samples (or fewer) in any given fine‐grained subgroup. Evaluating a downstream model on such a small test set yields wide confidence intervals and inaccurate evaluations that mask how a model truly performs on that subgroup. \textbf{Synthetic data, when conditioned on subgroup attributes, can overcome this limitation} by generating large, representative cohorts for fine‐grained intersectional groups. This enables precise subgroup evaluation with narrow confidence intervals, improving transparency and fairness. Ultimately, such use of synthetic data contributes to more \textbf{trustworthy and equitable AI systems}, where performance metrics truly reflect clinical reality across all patient populations.

\paragraph{\textbf{Main Contributions}}

This work focuses on using synthetic ICU time‐series data for robust and trustworthy model evaluation, both globally and across demographic subgroups. Our goal is to ensure that evaluation metrics computed on synthetic data faithfully reflect true performance on real ICU patients. To this end, we develop an enhanced generative framework and systematically compare it to existing models: TimeDiff, TimeAutoDiff, and HealthGen regarding their ability to produce synthetic data that serve as reliable test proxies.

\begin{enumerate}
    \item \textbf{Enhanced TimeAutoDiff for Evaluation.} We introduce an improved version of the latent diffusion model TimeAutoDiff (optimized for model evaluation purposes), augmented with two distribution-alignment losses (MMD and consistency regularization). These additions reduce the TRTS (Train on Real, Test on Synthetic) performance gap by over 70\% while maintaining TSTR (Train on Synthetic, Test on Real) stability.
    \item \textbf{Benchmarking Synthetic Hold-Outs.} We systematically compare Enhanced TimeAutoDiff, TimeDiff, TimeAutoDiff, and HealthGen to assess which generators produce synthetic hold-outs that most closely reproduce “real-on-real” (TRTR) evaluation metrics: which synthetic hold-outs can be most reliably trusted as proxies for real test data.
    \item \textbf{Subgroup-Level Evaluation.} We evaluate whether synthetic data can support fine-grained, demographically stratified analyses by generating large, conditionally sampled subgroups, especially where real test data are insufficient for reliable statistics.
    \item \textbf{Open-Source Release.} All code, model checkpoints, and subgroup-evaluation pipelines are publicly available at: \href{https://github.com/mahmoudibrahim98/icu-auto-diff}{\texttt{github.com/mahmoudibrahim98/icu-auto-diff}}.    
\end{enumerate}

The remainder of this paper is organized as follows. Section 2 reviews related work. Section 3 summarizes the used architectures. Section 4 details the methodology and the experimental setup .Section 5 presents the experimental results. Section 6 discusses implications and limitations. Finally, Section 7 concludes.

\section{Related Work}

Very few studies explicitly use synthetic data for model evaluation or for assessing performance across intersectional subgroups. A structured regression framework was introduced in  \cite{regression-evaluation}. It pools information across demographic intersections to produce stable, confidence-aware subgroup accuracy estimates. SYNG4ME \cite{syng4me} similarly leverages synthetic cohorts to yield more reliable subgroup metrics than small held-out real subsets. However, this work does not address time-series data or medical datasets.

Synthetic EHR generation must account for the inherent heterogeneity of real‐world ICU records, which combine static demographics, high‐cardinality categorical codes (e.g., ICD), and multivariate vital‐sign time series. Broadly, researchers distinguish four EHR formats: snapshot, aggregated, longitudinal, and time‐dependent, each presenting unique modeling challenges ~\cite{review}; our work focuses on the time‐dependent setting. Early efforts like TimeGAN ~\cite{TimeGAN} adapted the GAN framework to capture temporal dynamics, while EHR-M-GAN ~\cite{EHR-M-GAN} introduced separate encoders for each data modality to better synthesize mixed‐type sequences. More recently, HALO ~\cite{HALO} leveraged a hierarchical autoregressive language model to generate medical codes and visit sequences over long horizons, and HealthGen ~\cite{HealthGen} adopted a VAE to produce continuous ICU trajectories. Diffusion-based approaches have also emerged: Diffusion-TS ~\cite{Diffusion-TS}  uses a transformer-based latent diffusion to model complex time series, and TimeDiff ~\cite{TimeDiff} combines multinomial and Gaussian diffusion processes specifically for EHR data. Finally, FLEXGEN-EHR ~\cite{FLEXGEN-EHR} and TimeAutoDiff ~\cite{timeautodiff} employs latent diffusion to synthesize heterogeneous longitudinal records end-to-end. It is noted that there is a noticeable lack of emphasis on conditional generation and incorporating patient and demographic context in the generation process.

\section{Generative Models}

\subsection{Base Architecture}
The base architecture used in this work is TimeAutoDiff ~\cite{timeautodiff}, which consists of two main components:
\paragraph{Phase 1: $\beta$ -VAE   Training}
The autoencoder (consisting of GRU layers) maps mixed-type time-series features to a latent space $z$ using the loss:
$\mathcal{L}_{Auto} = \mathcal{L}_{recon} + \beta \mathcal{L}_{KLD}$ where $\mathcal{L}_{recon}$ combines MSE loss for continuous features and cross-entropy for discrete features, and $\mathcal{L}_{KLD} = \text{KL}(q(z|x) | p(z))$ enforces a Gaussian prior.

\paragraph{Phase 2: Conditional Diffusion Training}
A bidirectional RNN diffusion model operates on the latent space by corrupting and denoising $z$ using the loss function
$\mathcal{L}_{Diffusion} = \mathbb{E}_{t,\epsilon}[\|\epsilon - \epsilon_\theta(z_t, t, c)\|^2]$, 
where $c$ represents demographic conditioning information and classifier-free guidance balances condition adherence with diversity.


\subsection{Enhanced TimeAutoDiff: Our Contributions}
We introduce \textbf{Enhanced TimeAutoDiff}, an improved variant of the TimeAutoDiff generator that incorporates additional regularization to encourage a smoother and more structured latent space. Specifically, we augment each base loss with two auxiliary terms:

\paragraph{Maximum Mean Discrepancy (MMD) Loss:}

\[
\ell_{\mathrm{MMD}}\;=\;\frac{1}{B^2}\sum_{i,j}k(x_i,x_j)
+\frac{1}{B^2}\sum_{i,j}k(y_i,y_j)
-\frac{2}{B^2}\sum_{i,j}k(x_i,y_j),
\]
where $k(\cdot,\cdot)$ is a radial basis function kernel. This term enforces distributional alignment between the latent representations of real and synthetic samples.

\paragraph{Consistency Regularization:}
\[
\ell_{\mathrm{consistency}} = \mathrm{MSE}\bigl(f(x), f(x + \delta)\bigr),
\quad \delta \sim \mathcal{N}(0,\sigma^2I),
\]
This regularizer promotes local smoothness by constraining the model’s output to remain consistent under small Gaussian perturbations of the input, thereby improving the robustness of generated samples.

The overall training objectives become:
\[
\mathcal{L}_{\mathrm{Auto}}^{Enhanced}
=
\mathcal{L}_{\mathrm{Auto}}
+ \lambda_{Auto}^{MMD}\,\ell_{\mathrm{MMD}}
+ \lambda_{Auto}^{consistency}\,\ell_{\mathrm{consistency}},
\]
\[
\mathcal{L}_{\mathrm{Diffusion}}^{Enhanced}
=
\mathcal{L}_{\mathrm{Diffusion}}
+ \lambda_{diff}^{MMD}\,\ell_{\mathrm{MMD}}
+ \lambda_{diff}^{consistency}\,\ell_{\mathrm{consistency}}.
\]
The model trained under these enhanced objectives constitutes our \textbf{Enhanced TimeAutoDiff} framework.

\subsection{Baseline Models for Comparison}

In addition to the TimeAutoDiff baseline, we compare to two models: TimeDiff and HealthGen.
\textbf{TimeDiff} \cite{TimeDiff} is a diffusion probabilistic model that also uses BiRNN architecture for realistic privacy-preserving EHR time series generation. To simultaneously generate both continuous and discrete-valued time series, TimeDiff introduces a mixed diffusion approach that combines multinomial and Gaussian diffusion for EHR time series generation. The \textbf{HealthGen} \cite{HealthGen} model consists of a dynamical VAE-based architecture that allows
for the generation of feature time series with informative missing values, conditioned on high level static variables and binary labels.

\section{Methodology}

\begin{figure}[t]
\centering
\includegraphics[width=\textwidth]{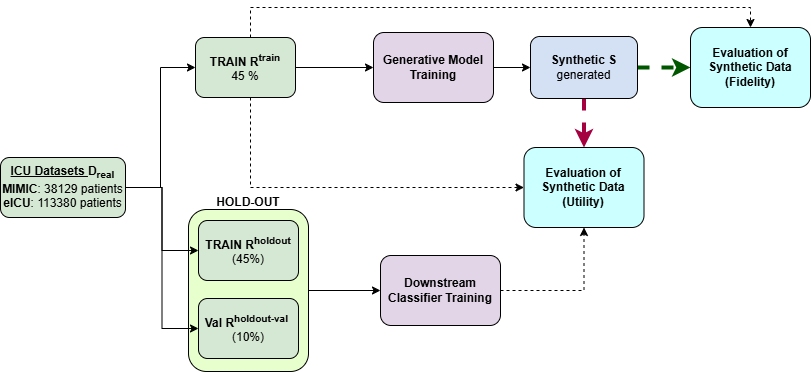}
\caption{\textbf{Workflow for Generative Model Training and Evaluation of Synthetic ICU Data.} The real ICU datasets (MIMIC and eICU) are split into training ($R^{train}$, 45\%), holdout ($R^{holdout}$, 45\%), and validation ($R^{holdout-val}$, 10\%) subsets. A generative model is trained on $R^{train}$ to produce synthetic data $S$. The synthetic data are assessed through two complementary evaluations: fidelity (green- intrinsic comparison with real data) and utility (purple- extrinsic evaluation via downstream classifier performance). Solid lines indicate training flows, while dashed lines denote evaluation processes.}
\label{overview}
\end{figure}

\subsection{Data}
Let \[
D_{\mathrm{Real}} = \{\, (x_i,\,y_i,\,c_i)\,\}_{i=1}^N
\] be the real ICU dataset of size $N$, where $x_i$ denotes the time-series data, $y_i$ the outcome , and $c_i$ the demographic attributes. As seen in fig.\@ \ref{overview}, we split $D_{Real}$ into disjoint subsets: a training set $R^{train}$ (45 \% of $|D_{real}|$), a holdout set $R^{holdout}$ (45 \%), and holdout validation set $R^{holdout-val}$ (10 \%). The generative model $G$ is trained on $R^{train}$, producing a synthetic dataset $S$ of equal size $N_{train}$ (possibly conditioned on y and c). Since $R^{train}$ serves both as the training source and reference for synthetic data evaluation, we omit the superscript hereafter and denote it simply as $R$. 

The synthetic data $S$ are evaluated through two complementary procedures: (i) \textit{fidelity}, assessing how closely $S$ matches the statistical properties of the real data $R$; and (ii)\textit{ }utility, evaluating how well $S$ preserves downstream predictive performance when used to train or evaluate classifiers. The holdout subsets $R^{holdout}$ and $R^{holdout-val}$ are used respectively for training and validating the downstream classifiers. This splitting strategy ensures sufficient data for both training a stable generative model and downstream model while maintaining strict separation between training and testing data. All splits are stratified by outcome and demographic variables to preserve population balance.

We use the publicly available MIMIC‐III ~\cite{mimic} and eICU~\cite{eicu} ICU EHRs, which consist of deidentified ICU data linked by patient ID and timestamp. Following established cohort‐selection criteria ~\cite{yaib}, we extract each patient’s first 24 hours of hourly measurements, along with static demographics (age, sex, ethnicity). We frame two popular downstream tasks on this 24-hour window: ICU mortality prediction ~\cite{yaib} and binary Length of Stays (LOS) (> 3 days), and consider 8 hourly features for the prediction: four vital signs (heart rate, respiratory rate, oxygen saturation, and mean blood pressure), plus four binary indicators for whether each value is missing. Missingness itself is retained via a mask, and missing values are forward‐filled. The missingness masks are used as input for the models. After preprocessing, eICU yields 113 380 patients (mortality = 5.51 \%, LOS > 3 d = 42.3 \%), and MIMIC yields 38 129 patients (mortality = 8.13 \%, LOS > 3 d = 48.7 \%).

\subsection{Synthetic Data Generation}

We adopt a standardized generation protocol across four architectures: Enhanced TimeAutoDiff, baseline TimeAutoDiff, HealthGen, and TimeDiff to ensure fair comparison. All models are trained on the same 45\% training split ($R_{\text{train}}$) with identical demographic conditioning on age group, gender, ethnicity, and outcome. Each generator produces a synthetic dataset of equal size ($|S| = |R_{\text{train}}|$), preserving the conditioning distribution of the real data.

Since TimeDiff is not inherently conditional, we adapt it by including demographic and outcome variables as additional input features during training. Conditional samples are then obtained via rejection sampling, retaining only those that match the target demographic and outcome attributes. Although less efficient than native conditional generation, this ensures that TimeDiff yields synthetic data with population characteristics consistent with the conditional models.

\subsection{Downstream Prediction Models}

We use a one-layer Gated Recurrent Unit (GRU) network followed by a fully connected sigmoid output layer. Models are trained with binary cross-entropy loss using the Adam optimizer (learning rate = $5\times10^{-4}$, batch size = 64) for up to 50 epochs, selecting the best checkpoint based on validation performance. Architectures and hyperparameters are fixed across all experiments for consistency and follow standard practices from the literature \cite{HealthGen,yaib}. Ablation studies confirm that downstream model parameters have minimal influence on the relative performance of synthetic data generators, indicating that our findings reflect differences in synthetic data quality rather than downstream model optimization.

We implement a one‐layer Gated Recurrent Unit  (GRU) network followed by a fully connected layer with sigmoid output. Training is done using binary cross‐entropy loss, an Adam optimizer (lr = $5\times10^{-4}$), a batch size = 64. We fix these architectures across all experiments for consistency. In all evaluations, we train each downstream model for up to 50 epochs, selecting the best checkpoint via validation performance. We used standard hyperparameters and model architectures from the literature \cite{HealthGen,yaib}. Our ablation studies demonstrate that downstream model parameters have minimal effect on the relative performance comparisons between synthetic data generators, ensuring that our conclusions are robust to these choices. The focus of our evaluation is on comparing synthetic data quality rather than optimizing downstream model performance.

\subsection{Evaluation Metrics}
Our goal is twofold: (1) to improve global evaluation utility (TRTS) so that real‐trained models evaluated on synthetic data match performance on true held‐out real data, and (2) to enable accurate subgroup‐level evaluation by generating large synthetic cohorts conditioned on demographic labels.

\paragraph{\textbf{Utility for Downstream Training}}
\begin{figure}[t]
\centering
\includegraphics[width=0.8\textwidth]{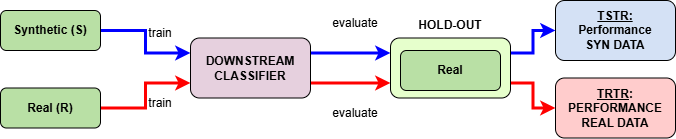}
\caption{Training Utility workflow.}
\label{training-utility}
\end{figure}
Using identical downstream architectures and hyper-parameters, we train two downstream models, one on real data (red path) and another on synthetic data (blue path). Then, we evaluate both models on hold-out real data $R^{holdout}$ as seen in fig.\@ \ref{training-utility}. $\text{TRTR}_{\text{train}}$ and $\text{TSTR}_{\text{train}}$ refer to "Train on Real, Test on Real" and "Train on Synthetic, Test on Real," respectively, and are used to evaluate the training utility of synthetic data.
This evaluation is repeated 5 times, each time on a differently trained downstream model.


\paragraph{\textbf{Utility for Downstream Evaluation}}
\begin{figure}[t]
\centering
\includegraphics[width=0.8\textwidth]{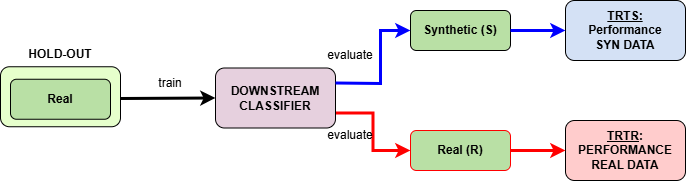}
\caption{Evaluation Utility workflow.}
\label{evaluation-utility}
\end{figure}
We train a downstream predictive model (mortality classifier or LOS classifier) on real holdout dataset $M_{\mathrm{real}}=\mathrm{train}\left(R^{holdout}\right)$. Then, we evaluate on both synthetic (blue) and real (red) datasets as seen in fig.\@ \ref{evaluation-utility}.
$\text{TRTR}_{\text{evaluate}}$ and $\text{TRTS}_{\text{evaluate}}$ refer to "Train on Real, Test on Real" and "Train on Real, Test on Synthetic," respectively, and are used to evaluate the evaluation utility of synthetic data. This evaluation is repeated 5 times, each time on a differently trained downstream model.

\paragraph{\textbf{Utility for Subgroup Level Evaluation}}

We define three categorical demographic dimensions: age bracket ($<30$, $31$-$50$, $51$-$70$, $>70$), sex (M and F), and ethnicity (White, Black, Asian, Other). This yields $4 \times 2 \times 4 = 32$ intersectional subgroups of age $\times$ sex $\times$ ethnicity.

\begin{figure}[t]
\centering
\includegraphics[width=\textwidth]{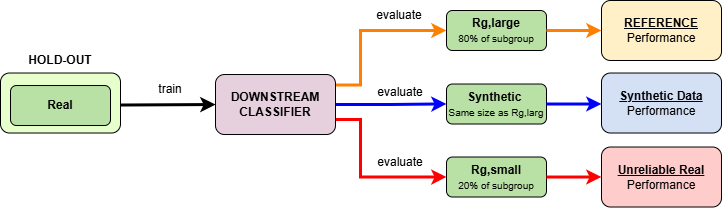}
\caption{Subgroup Utility workflow.}
\label{subgroup-utility}
\end{figure}
For each demographic subgroup $g$, let $R_g \subset R_{\text{train}}$ be real samples with subgroup label $g$. To establish reliable reference evaluations, we allocate 80\% of each demographic bucket as the ``ground truth'' set $R_{g,\text{large}}$. The downstream model is evaluated on this set to obtain stable performance estimates. The remaining 20\% ($R_{g,\text{small}}$) simulates the small test samples typically available for minority subgroups in real datasets. The 80/20 split is repeated 5 times with different random seeds to ensure statistical robustness.

We generate synthetic cohort $S_g$ with $|S_g| = |R_{g,\text{large}}|$ and compute:

\begin{align}
\epsilon_g^{\text{naive}} &= |\text{AUROC}(M_{\text{real}}, R_{g,\text{large}}) - \text{AUROC}(M_{\text{real}}, R_{g,\text{small}})| \\
\epsilon_g^{\text{synth}} &= |\text{AUROC}(M_{\text{real}}, R_{g,\text{large}}) - \text{AUROC}(M_{\text{real}}, S_g)|
\end{align}
where $\epsilon_g^{\text{naive}}$ represents the error when using small real test sets, and $\epsilon_g^{\text{synth}}$ represents the error when using large synthetic cohorts, both compared to the ground truth performance on $R_{g,\text{large}}$.



\paragraph{\textbf{Discriminative Fidelity Score}}
We split the real and synthetic datasets into train and test subsets $R_{train}$, $R_{test}$ and $S_{train}$, $S_{test}$ respectively. We then label every example in \(R_{\text{train}}\) with \(0\) (real), and every example in \(S_{\text{train}}\) with \(1\) (synthetic). We train a discriminator on this balanced dataset containing \(R_{\text{train}}\) and \(S_{\text{train}}\). Let the discriminator assign a score \(p(\text{synthetic})\) to each sample in the combined dataset ($R_{\text{test}} \cup S_{\text{test}}$).
\[
  \mathrm{DiscAUC}_{S}
    \;=\;
  \mathrm{AUC}\bigl[y \in \{0,1\},\,p(\text{synthetic})\bigr].
\]

\subsection{Experimental Repetition}
We generate 5 independent synthetic datasets from each generator and train 5 downstream models per dataset (real or synthetic), yielding 25 evaluation runs per experimental condition. Moreover, the process of splitting the subgroups into small and large groups is repeated 5 times. This accounts for variability in both synthetic data generation, downstream model training and evaluation, and data splits, enabling robust statistical analysis with reliable confidence intervals and significance testing, while balancing between statistical rigor and computational feasibility. All reported metrics include 95\% CIs based on the independent runs.

\subsection{Alignment weights search}
We explored nine configurations that systematically vary the four alignment weights $\lambda_{AE\text{-}MMD}$,$\lambda_{AE\text{-}consist}$,$\lambda_{diff\text{-}MMD}$,$\lambda_{diff\text{-}consist}$
across three regimes (zero, light = 0.1, moderate = 0.5). In practice, each experiment requires full diffusion training (on eICU or MIMIC), synthetic sample generation, and downstream training/evaluation. Therefore, this choice of weights and number of configurations ensures sufficient coverage of “no regularization,” “light touch,” and “moderate emphasis” within a timely possible completion. Moreover, this choice helps isolate individual effects and probe pairwise combinations, avoiding a combinatorial explosion of configurations



\section{Experimental Results}
We present our results in three parts. First, we examine the discriminative fidelity score on eICU and MIMIC for both mortality and LOS tasks, then we present global evaluation utility (TSTR vs TRTS). Finally, we examine subgroup‐level evaluation, comparing small real test sets to pooled real “ground truth” and to large synthetic cohorts conditioned on subgroup labels.

\subsection{Discriminative Fidelity Comparison}

\begin{figure}[t]
\centering

\includegraphics[width=0.6\textwidth]{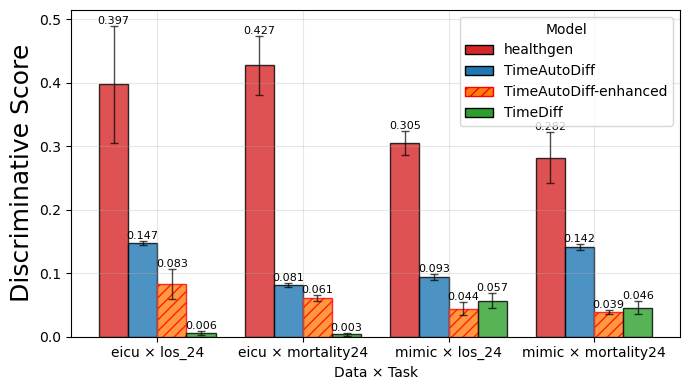}

\caption{\textbf{Model Comparison - Discriminative Fidelity} This figure displays bar charts comparing the Discriminative Score of HealthGen, TimeAutoDiff, Enhanced TimeAutoDiff and TimeDiff across different tasks and datasets. The lower the scores, the better. } \label{fig2}
\end{figure}

Fig.\@ \ref{fig2} presents discriminative AUCs for four synthetic data generators: HealthGen (conditional VAE), TimeAutoDiff (autoencoder + diffusion), Enhanced TimeAutoDiff (TimeAutoDiff with extra penalty), and TimeDiff (pure diffusion) on four Data × Task combinations (e.g. eICU $LOS_{24}$, MIMIC $LOS_{24}$, eICU $Mortality_{24}$, MIMIC $Mortality_{24}$).

HealthGen’s discriminative scores range from 0.282 to 0.427, meaning a simple classifier can easily distinguish its outputs from real data. TimeAutoDiff improves on this with discriminative AUCs of 0.081–0.147, while TimeDiff performs best, with AUCs as low as 0.003–0.057.
Across every setting, pure diffusion (TimeDiff) is hardest to distinguish from real, followed closely by the Enhanced TimeAutoDiff. The original TimeAutoDiff is an order of magnitude much better than HealthGen, but the added alignment losses yield a substantial drop (25-70\%) in the discriminative score compared to the baseline.

\subsection{Global Utility Comparison}
\begin{figure}[t]
\centering
\includegraphics[width=0.8\textwidth]{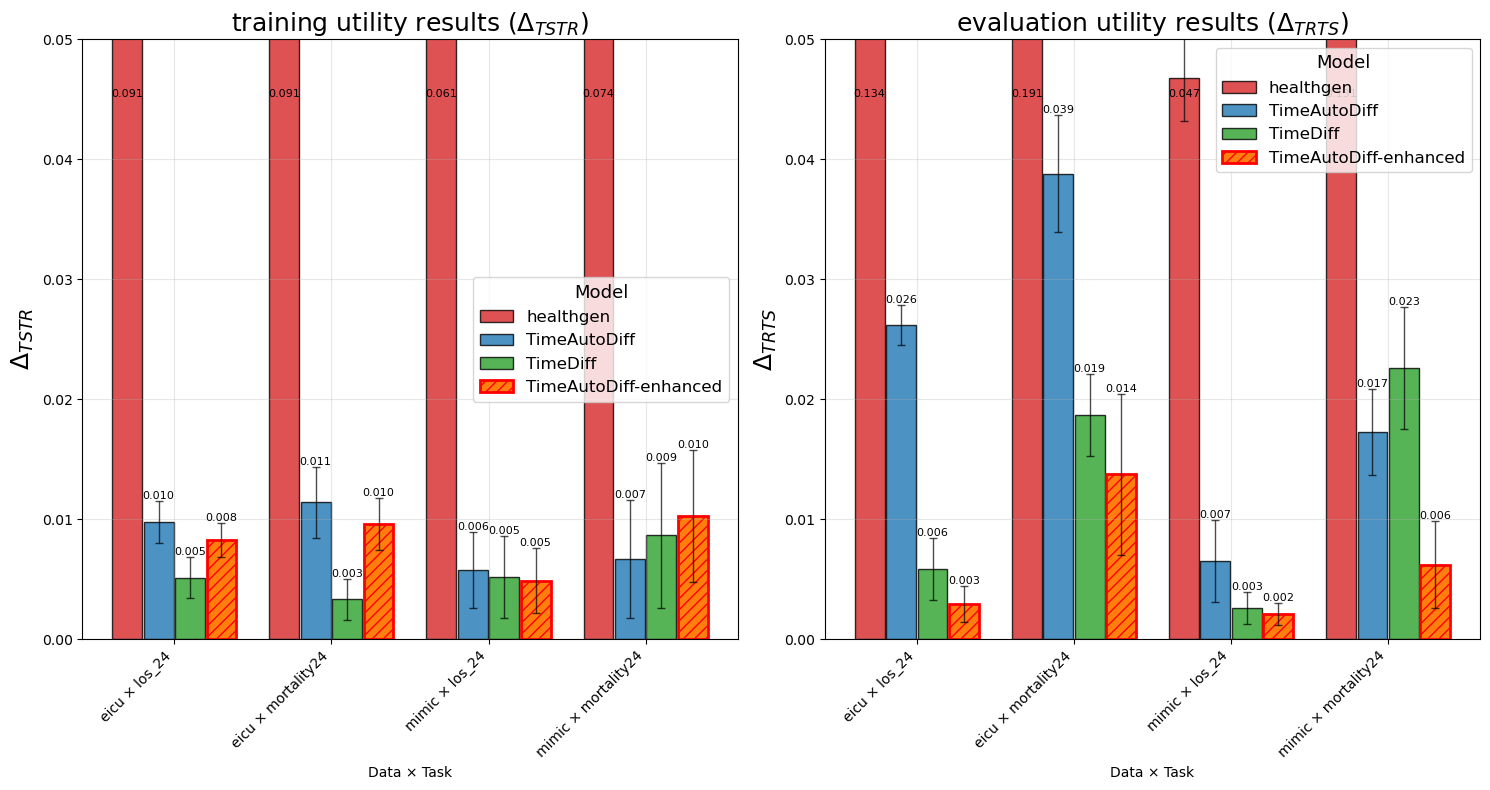}
\caption{\textbf{Model Comparison- Global Downstream Utility } This figure displays bar charts comparing the downstream utility for training ($\Delta_{TSTR}$, left panel) and evaluating ($\Delta_{TRTS}$, right panel) of HealthGen,TimeAutoDiff, Enhanced TimeAutoDiff, and TimeDiff across different tasks and datasets.
The lower the scores, the better. 
For clarity, the y-axis is limited to 0.05, truncating larger values (e.g., HealthGen) to highlight differences among the other models.} \label{fig1}
\end{figure}

We then evaluated the global downstream utility of the four synthetic data generators. As shown in the left‐hand panel of fig.\@ \ref{fig1}, HealthGen exhibits large $\Delta_{TSTR}$ gaps ($\approx$ 0.06-0.10), indicating that downstream models trained on HealthGen synthetic data perform substantially worse on real test data than fully real trained baselines. Both TimeAutoDiff and the enhanced variant exhibit similar performance and reduce $\Delta_{TSTR}$ to $\approx$ 0.006–0.011, and TimeDiff further drives it down to $\approx$ 0.003–0.009 across all tasks.

In the right‐hand panel, HealthGen again performs worst with  $\Delta_{TRTS}$ gaps ($\approx$ 0.13–0.19). TimeAutoDiff cuts those evaluation gaps by roughly 75 \% (to $\approx$ 0.017–0.039), while TimeDiff drives it down to $\approx$ 0.003–0.023. The Enhanced TimeAutoDiff delivers the smallest evaluation gaps of all four: between 0.003 and 0.014 AUROC depending on the task. For instance, on eICU $LOS_{24}$ it achieves just 0.003, and on MIMIC $Mortality_{24}$ 0.006. This shows that our regularized variant of TimeAutoDiff yields the most faithful synthetic hold‐outs for downstream evaluation, and when compared to the left-hand panel, this was done at a negligible cost to the generator’s ability to produce useful training data.

Comparing the TimeAutoDiff and our enhanced variant of it, we see that the “train on real, test on synthetic” gap shrinks dramatically under the best‐enhanced setting. For example, eICU $LOS_{24}$ drops from a gap of 0.026 to 0.003, and MIMIC $Mortality_{24}$ falls from 0.017 to 0.006. In each case, the enhanced version (model trained with the optimized weights) reduces $\Delta_{TRTS}$ by roughly 70–90\%. This means that the synthetic data can give a nearly identical performance as held-out real data for a trained model, all at almost zero cost to the generator’s ability to produce synthetic data that trains a high‐quality downstream model, as the $\Delta_{TSTR}$ remains very small ($\approx$ 0.01) both before and after enhancement. Therefore, by adding MMD and consistency losses to TimeAutoDiff and by tuning their weights per Data $x$ Task, we significantly reduce the evaluation gap $\Delta_{TRTS}$ without sacrificing $\Delta_{TSTR}$  or fidelity.

\begin{table}[t]
  \centering            
\caption{\textbf{Comparison of Downstream Evaluation and Downstream Training.} This table illustrates some examples where the difference between training utility ($\Delta_{TSTR}$) and evaluation utility ($\Delta_{TRTS}$) across various tasks and datasets is significantly big.}\label{tab1}
\begin{tabular}{cccccc} 
\toprule
Data &  Task & Model & $\Delta_{TSTR}$  & $\Delta_{TRTS}$ & Relation  \\
\midrule
eICU &  $LOS_{24}$ & TimeAutoDiff & $0.01 \pm 0.002$ & $0.026 \pm 0.002$ &  $\Delta_{TRTS} \approx 3 \times \Delta_{TSTR}$    \\
eICU &  $Mortality_{24}$ & TimeAutoDiff & $0.011 \pm 0.003$ & $0.039 \pm 0.005$ & $\Delta_{TRTS} \approx 3.5 \times \Delta_{TSTR}$   \\
eICU &  $Mortality_{24}$ & TimeDiff & $0.003 \pm 0.002$ & $0.019 \pm 0.003$ &  $\Delta_{TRTS} \approx 6 \times \Delta_{TSTR}$   \\
\bottomrule
\end{tabular}
\end{table}

Table.\@ \ref{tab1} and fig.\@ \ref{fig1} together show that, even though a downstream model trained on synthetic data (TSTR) almost matches the performance of a model trained on real data,  the reverse (training on real and testing on synthetic TRTS) shows a larger discrepancy. For example, looking at the data generated by TimeDiff, specifically for eICU $Mortality_{24}$,  $\Delta_{TSTR}$ $\approx$ 0.003 while $\Delta_{TRTS}$ is $\approx$ 0.019, 6 times larger. This problem was substantially mitigated in the Enhanced TimeAutoDiff.


\subsection{Subgroup‐Level Evaluation}

\begin{table}[]
  \centering            
\caption{Percentage of intersectional subgroups in which the synthetic evaluation error is strictly lower than the real‐test error, for each generator:TimeAutoDiff (TA), TimeDiff (TD), and Enhanced TimeAutoDiff $TA^{+}$ across the different configurations.
}\label{tab2}

\begin{tabular}{ccccc}
\toprule
\multirow{2}{*}{Data} & \multirow{2}{*}{Task} & \multicolumn{3}{l}{\% of Subgroups Where Synthetic $<$ Test Error} \\
 &  & TA & TD & $TA^{+}$ \\ 
 \midrule

eICU & $Mortality_{24}$ & 48\% & 60\% & 76\% \\
eICU & $LOS_{24}$ & 44\% & 52\% & 72\% \\
mimic & $Mortality_{24}$ & 68\% & 68\% & 84\% \\
mimic & $LOS_{24}$ & 72\% & 56\% & 76\% \\ 
\bottomrule
\end{tabular}
\end{table}

\begin{figure}[t]
\centering
\includegraphics[width=0.8\textwidth]{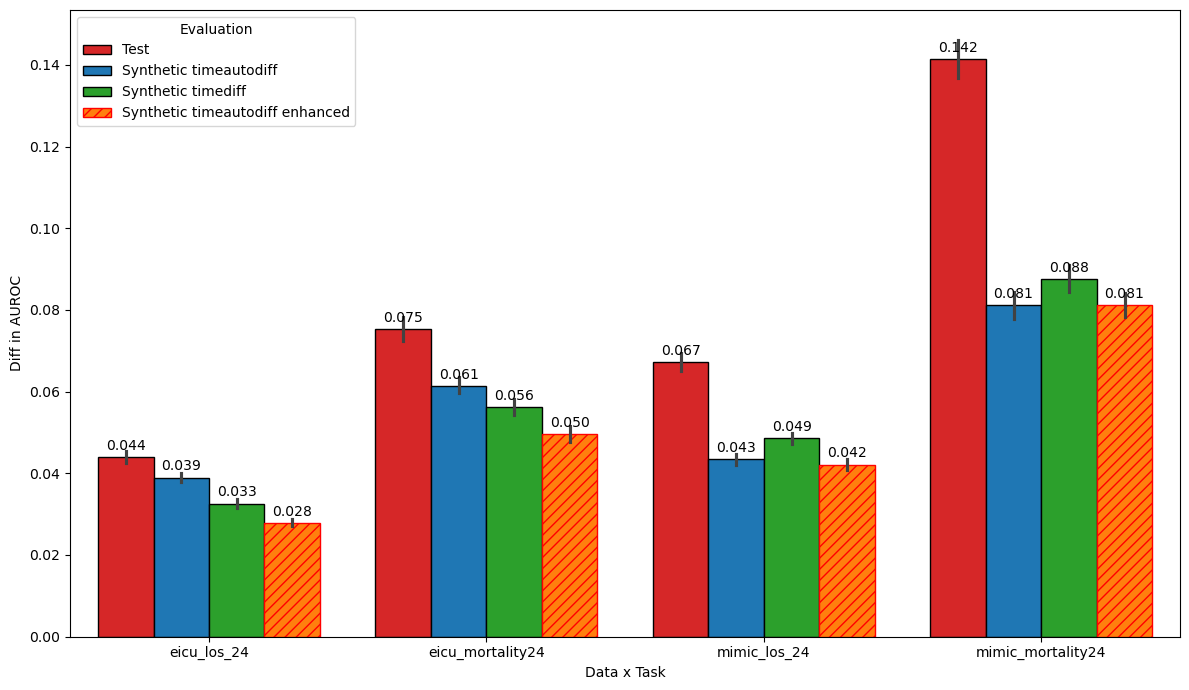}
\caption{\textbf{Subgroup-Level Evaluation} Mean absolute AUROC error of small‐real vs. large‐pool (“Test,” red) compared to synthetic cohort vs. large‐pool errors using TimeAutoDiff (blue) and TimeDiff(green), and Enhanced TimeAutoDiff (dashed orange)  averaged over all subgroups for different data and task configurations. The lower the scores, the better.} \label{fig4}
\end{figure}

For each Data $x$ Task,  we compute the absolute AUROC distance between (a) a small real‐test subgroup and the large “ground‐truth” subgroup (red bars labeled “Test”) versus (b) a synthetic, conditional‐on‐subgroup cohort and that same ground‐truth subgroup (orange,green, and blue bars labeled “Synthetic”). 
Then, we average those errors across all subgroups to produce the red (“Test”), blue (“Synthetic TimeAutoDiff”), green ("Synthetic TimeDiff"), and dashed orange  (“Enhanced TimeAutoDiff”) bars in Fig.\@ \ref{fig4}. In Table~\ref{tab2}, we counted, across all subgroups, how often the Synthetic error was significantly smaller than the Test error.

Fig.\@ \ref{fig4} shows that, in every case, the synthetic cohorts cut the error roughly in half compared to the small real‐test subsets (red bar), and our enhanced generator (Enhanced TimeAutoDiff) yields the lowest errors of all. For example, on eICU $LOS_{24}$ the mean error drops from 0.044 (Test) to 0.039 (TimeAutoDiff), 0.033 (TimeDiff), and 0.028 (Enhanced TimeAutoDiff); on MIMIC $Mortality_{24}$ it falls from 0.142 to 0.081, 0.088, and 0.081 respectively; On eICU $Mortality_{24}$: 0.075 → 0.061 → 0.056 → 0.050, and on MIMIC $LOS_{24}$: 0.067 → 0.043 → 0.049 → 0.042.

Table~\ref{tab2} shows the fraction of subgroups in which each synthetic cohort outperforms the real‐test subset (i.e. has a strictly smaller error). Although both TimeAutoDiff and TimeDiff reduce the mean Test error substantially as depicted in \ref{fig4}, almost halving it in most cases, their success rates remain under 50 \% on the eICU tasks (TimeAutoDiff: 48 \% for $Mortality_{24}$, 44 \% for $LOS_{24}$; TimeDiff: 60 \% for $Mortality_{24}$  and 52 \% for $LOS_{24}$), meaning they only outperform the real‐test evaluation in fewer than half of the subgroups. By contrast, our Enhanced TimeAutoDiff model outperforms Test in the majority of subgroups across every setting (72 \%–84 \% success). Moreover, the Enhanced TimeAutoDiff achieves the lowest average errors in every Data × Task.

Together, these results demonstrate that generating large conditional cohorts, even with different underlying generator architectures, yields substantially more accurate subgroup performance estimates than evaluating directly on small held-out subsets, and that our Enhanced TimeAutoDiff model turns synthetic cohorts into more reliable evaluators that deliver more accurate subgroup performance estimates in the vast majority of cases.

\section{Discussion}

This study shifts focus from synthetic data as a training resource to synthetic data as an evaluation tool, investigating how well synthetic ICU time-series can serve as reliable proxies for real data in model assessment. We systematically compared four generators across global and subgroup-level evaluation scenarios, with particular emphasis on the previously understudied evaluation utility (TRTS) gap.

\subsection{Key Findings and Implications}

Our experiments reveal three critical insights about synthetic medical data evaluation. First, \textbf{diffusion-based models substantially outperform VAE-based approaches} not only in terms of evaluation, but also in terms of training utility and discriminative fidelity. While HealthGen exhibits large evaluation gaps ($\Delta_{TRTS}$ = 0.13–0.19 AUROC), diffusion models reduce these gaps to 0.003–0.04 AUROC, with Enhanced TimeAutoDiff achieving the smallest gaps (0.003–0.014 AUROC). In terms of discriminative fidelity, pure diffusion (TimeDiff) achieved the lowest discriminative scores (AUC $\approx$ 0.003–0.057), with Enhanced TimeAutoDiff close behind (AUC $\approx$ 0.039–0.083), followed by TimeAutoDiff (0.081–0.147), while HealthGen remained easiest to distinguish from real data (AUC $\approx$ 0.28–0.43).

Second, \textbf{evaluation utility requires different optimization than training utility}. Although diffusion models achieve strong TSTR performance ($\Delta_{TSTR} \approx$ 0.01), they initially showed substantially larger TRTS gaps, revealing fundamental distributional mismatches that standard training objectives fail to address.

Third, \textbf{synthetic cohorts enable reliable subgroup evaluation where real data are insufficient}. Enhanced TimeAutoDiff outperforms small real test sets in 72–84\% of demographic subgroups, reducing evaluation error by up to 50\% compared to standard approaches that rely on limited real samples for minority populations.

\subsection{Technical Innovation: Distribution Alignment}

Although diffusion‐based generators yield strong TSTR performance, they still exhibit substantially larger than TRTS gaps, revealing residual mismatches when evaluating real‐trained models on synthetic hold‐outs. This pronounced TSTR-TRTS gap reflects an inherent limitation of generative models: they tend to produce smoothed versions of training data, averaging out rare modes while preserving dominant patterns. During training (TSTR), downstream models can adapt to these smoothed distributions. However, during evaluation (TRTS), fixed models encounter mismatches between their learned decision boundaries and the synthetic distribution, leading to evaluation errors.

Enhanced TimeAutoDiff addresses this through explicit distribution alignment. We augment both autoencoder and diffusion losses with MMD penalties that enforce distributional matching and consistency losses that ensure smooth model behavior. This approach reduces $\Delta_{TRTS}$ by 70–90\% (Across both datasets and both tasks, $\Delta_{TRTS}$ fell to 0.003–0.014 AUROC, outperforming all baselines) while preserving training utility ($\approx$ 0.01 AUROC), demonstrating that evaluation-specific optimization can substantially improve synthetic data reliability without sacrificing sample quality.

Moreover, the enhanced approach demonstrated improved subgroup evaluation. Our enhanced model outperforms the real test evaluation in 72–84 \% of all subgroups, and yields the lowest mean errors in every Data$\times$Task. This confirms that generating large, conditionally matched synthetic cohorts using our enhanced model is not only beneficial on average, but reliably so for intersectional groups where real data are scarce. In contrast, baseline models (TimeAutoDiff and TimeDiff) only succeeded in fewer than 50 \% of eICU subgroups, meaning that in the majority of subgroups, real test subsets still outperform their synthetic counterparts.

\subsection{Clinical Impact and Fairness Implications}

Our subgroup evaluation results have direct implications for healthcare AI fairness. Traditional evaluation approaches often rely on small subgroup samples that yield unreliable performance estimates, masking true model behavior across demographic intersections. By generating large, balanced synthetic cohorts, we enable precise bias detection and fair performance assessment, facilitating more equitable AI systems while maintaining patient privacy.

\subsection{Limitations and Future Directions}

Several limitations warrant acknowledgment. Our hyperparameter optimization remains dataset-specific, requiring manual tuning for optimal performance. \textbf{Meta-learning approaches} could automate this process, improving robustness across different medical contexts.\textbf{Privacy guarantees} represent another critical gap. While MMD and consistency losses implicitly discourage memorization, formal differential privacy analysis is essential for deployment in sensitive healthcare settings. Incorporating DP-SGD training and conducting rigorous privacy audits should be immediate priorities. Finally, our evaluation focuses on two prediction tasks (mortality and LOS) within ICU settings. \textbf{Broader medical applications} including continuous outcomes, multimodal data, and other clinical domains require investigation to establish the generalizability of our approach.



\section{Conclusion}
We have introduced Enhanced TimeAutoDiff, a latent diffusion framework augmented with distribution‐alignment losses, and demonstrated its superiority for both global and subgroup‐level model evaluation on ICU time‐series data. Our extensive benchmarks on MIMIC‐III and eICU show that Enhanced TimeAutoDiff achieves the smallest evaluation gaps ($\Delta_{TRTS}$ $\leq$ 0.014 AUROC) and preserves training utility ($\Delta_{TSTR} \approx$  0.01). By generating large conditional cohorts, it also halved subgroup‐level estimation errors and outperformed small real test sets in the majority of intersectional subgroups. These results establish a practical blueprint for using synthetic data to perform robust, privacy‐preserving evaluations of predictive models across diverse patient populations. All code, checkpoints, and evaluation pipelines are publicly available to facilitate adoption and further research.

\begin{credits}
\subsubsection{\ackname} This work was funded by the European Union under the Horizon Europe grant 101095435. Views and opinions expressed are however those of the author(s) only and do not necessarily reflect those of the European Union. Neither the European Union nor the granting authority can be held responsible for them.

\subsubsection{\discintname}
The authors have no competing interests to declare that are relevant
to the content of this article.
\end{credits}
%
%
%
%

\end{document}